\documentclass[10pt,twocolumn,letterpaper]{article}

\usepackage{ijcb}
\usepackage{times}
\usepackage{epsfig}
\usepackage{graphicx}
\usepackage[fleqn]{amsmath}
\usepackage{amssymb}
\usepackage{tabularx}
\usepackage{makecell}

\usepackage{float}
\usepackage{subfig}
\usepackage[flushleft]{threeparttable}
\usepackage{adjustbox}
\usepackage{tablefootnote}
\usepackage[font={footnotesize}]{caption}
\usepackage{tablefootnote}
\usepackage{amsmath}
\usepackage{xcolor} 
\usepackage{dblfloatfix} 
\usepackage{caption}
\usepackage{booktabs,array}
\usepackage[detect-none]{siunitx}
\usepackage{multicol}
\usepackage[usestackEOL]{stackengine}

\usepackage{algorithm}
\usepackage[noend]{algpseudocode}

\makeatletter
\def\algbackskip{\hskip-\ALG@thistlm}
\makeatother

\usepackage[pagebackref=true,breaklinks=true,colorlinks,bookmarks=false]{hyperref}

\ijcbfinalcopy 

\ifijcbfinal\fi
\begin{document}

\newcolumntype{C}{>{\centering\arraybackslash}p}

\title{Fingerprint Presentation Attack Detection:\\ A Sensor and Material Agnostic Approach}

\author{Steven A. Grosz\\
Michigan State University\\
East Lansing, MI, 48824\\
{\tt\small groszste@msu.edu}
\and
Tarang Chugh\\
Michigan State University\\
East Lansing, MI, 48824\\
{\tt\small chughtar@msu.edu}
\and
Anil K.~Jain\\
Michigan State University\\
East Lansing, MI, 48824\\
{\tt\small jain@msu.edu}
}

\maketitle
\thispagestyle{empty}

\begin{abstract}
   The vulnerability of automated fingerprint recognition systems to presentation attacks (PA), i.e., spoof or altered fingers, has been a growing concern, warranting the development of accurate and efficient presentation attack detection (PAD) methods. However, one major limitation of the existing PAD solutions is their poor generalization to new PA materials and fingerprint sensors, not used in training. In this study, we propose a robust PAD solution with improved cross-material and cross-sensor generalization. Specifically, we build on top of any CNN-based architecture trained for fingerprint spoof detection combined with cross-material spoof generalization using a style transfer network wrapper. We also incorporate adversarial representation learning (ARL) in deep neural networks (DNN) to learn sensor and material invariant representations for PAD. Experimental results on LivDet 2015 and 2017 public domain datasets exhibit the effectiveness of the proposed approach.
\end{abstract}

\section{Introduction}
    Fingerprints are considered one of the most reliable biometric traits due to their inherent uniqueness and persistence, which has led to their widespread adoption in secure authentication systems~\cite{maltoni2009handbook}. However, it has been demonstrated that these systems are vulnerable to presentation attacks by adversaries trying to gain access to the system~\cite{evans2019handbook, ODNI}. A presentation attack (PA) as defined by the ISO standard \textit{IEC 30107-1:2016(E)}~\cite{ISO-PA} is a ``\textit{presentation to the biometric data capture subsystem with the goal of interfering with the operation of the biometric system}.'' These attacks often involve a fingerprint cast from a mold using common household materials (gelatin, silicone, wood glue, etc) and aim to mimic the ridge-valley structure of an enrolled user's fingerprint~\cite{matsumoto2002impact, cao2016hacking, engelsma2018universal, yoon2012altered, marasco2014survey}.
    
    The vulnerability of these systems to presentation attacks led to a series of standard assessments of fingerprint presentation attack detection (PAD) methods developed by different organizations\footnote{In the literature, presentation attack detection (PAD) is also commonly referred to as spoof detection and liveness detection. In this work, we use these terms interchangeably.}. The First International Fingerprint Liveness Detection Competition debuted in 2009~\cite{marcialis2009first} with subsequent competitions every two years, the most recent being 2019~\cite{yambay2012livdet, ghiani2013livdet, mura2015livdet, mura2018livdet, orr2019livdet}. 
    
    \begin{figure}[t]
        \centering
        \includegraphics[width=\linewidth]{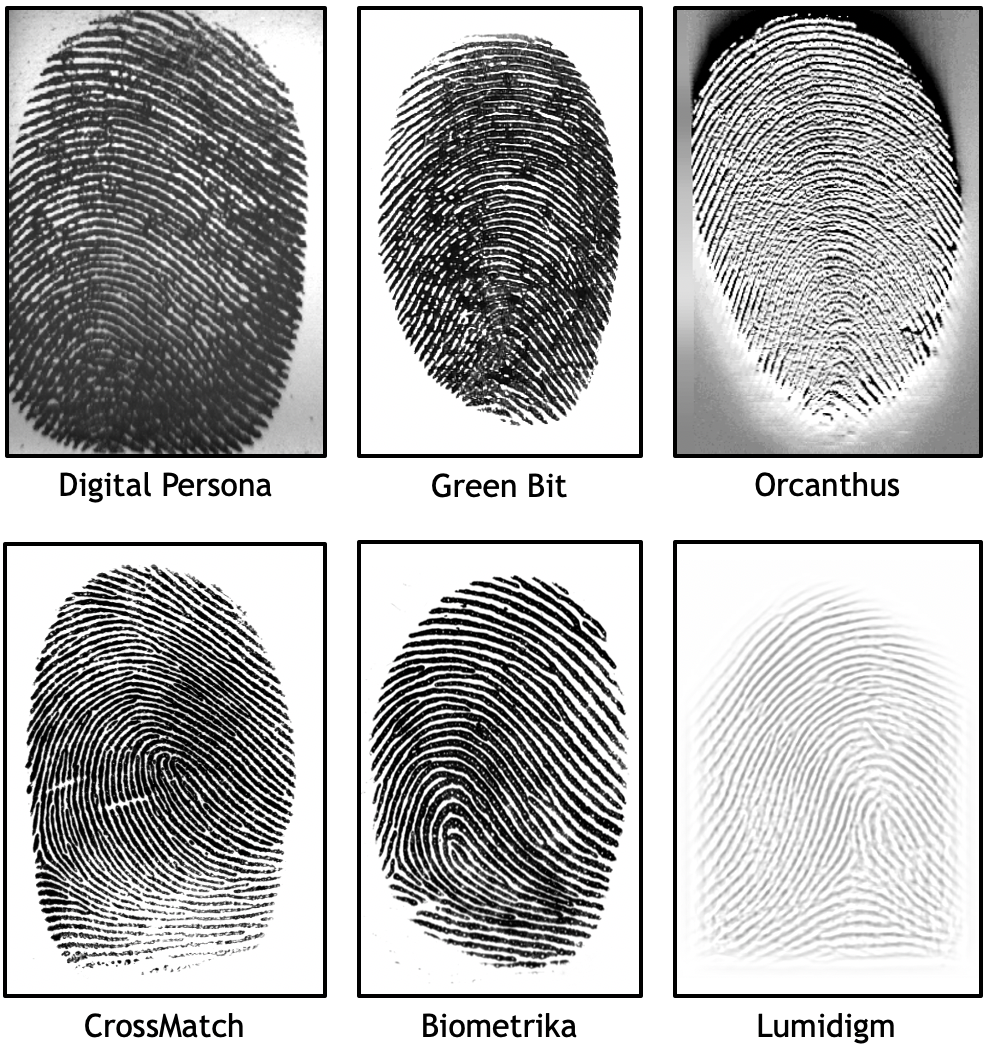}%
        \caption{Illustration of the differences in textural appearance of live fingerprints captured on six different fingerprint readers. Images from LivDet 2015~\cite{mura2015livdet}, LivDet 2017~\cite{mura2018livdet}, and MSU-FPAD datasets~\cite{chugh2018fingerprint}.}
        \label{fig:6_sensors}
        \vspace{-1.6em}
    \end{figure}
    
    
    \begin{table*}[t]
        \renewcommand{\arraystretch}{1.3}
        \caption{Summary of Published Fingerprint Cross-Material Generalization Studies.}
        \label{tab:prior_work}
        \small
        {
        \begin{tabular}{>{\centering\arraybackslash}p{0.16\textwidth}>{\centering\arraybackslash}p{0.31\textwidth}>{\centering\arraybackslash}p{0.14\textwidth}>{\centering\arraybackslash}p{0.29\textwidth}}
        Study & Approach & Database & Performance\\
        \toprule
        Rattani et al.~\cite{rattani2015open} & Weibull-calibrated SVM & LivDet 2011 & EER = 19.70 \%\\
        \hline
        Ding \& Ross~\cite{ding2016ensemble} & Ensemble of multiple one-class SVMs & LivDet 2011 & EER = 17.06 \%\\
        \hline
        Chugh \& Jain~\cite{chugh2018fingerprint} & MobileNet-v1 trained on minutiae-centered local patches & LivDet 2011-2015 & ACE = 1.48 \% (LivDet 2015), \newline 2.93 \% (LivDet 2011, 2013)\\
        \hline
        Chugh \& Jain~\cite{chugh2018generalization} & Identify a representative set of spoof materials to cover the deep feature space & MSU-FPAD v2.0, 12 spoof materials & TDR = 75.24 \% @ FDR = 0.2 \%\\
        \hline
        Engelsma \& Jain~\cite{engelsma2019generalizing} & Ensemble of generative adversarial networks (GANs) & Custom database with live and 12 spoof materials & TDR = 49.80 \% @ FDR = 0.2 \%\\
        \hline
        Gonzalez-Soler et al.~\cite{gonzalez2019fingerprint} & Feature encoding of dense-SIFT features & LivDet 2011-2015 & TDR = 7.03 \% @ FDR = 1.0 \% \newline (LivDet 2015), ACE = 1.01 \% \newline (LivDet 2011, 2013)\\
        \hline
        Tolosana et al.~\cite{tolosana2019biometric} & Fusion of two CNN architectures trained on SWIR images & Custom database with live and 8 spoof materials & EER = 1.35 \%\\
        \hline
        Gajawada et al.~\cite{gajawada2019universal} & Style transfer from spoof to live images with a few samples of target material & LivDet 2015, CrossMatch sensor & TDR = 78.04 \% @ FDR = 0.1 \%\\
        \hline
        Chugh \& Jain~\cite{chugh2019fingerprint} & Style transfer between known spoof materials to improve generalizability against unknown materials & MSU-FPAD v2.0, 12 spoof materials \& LivDet 2017 & TDR = 91.78 \% @ FDR = 0.2 \% (MSU-FPAD v2.0); Avg. Accuracy = 95.88 \% (LivDet 2017)\\
        \hline
        \textbf{Proposed Approach} & Style transfer with a few samples of target sensor fingerprint images + ARL & LivDet 2015 & TDR = 87.86 \% @ FDR = 0.2 \% cross-sensor \& cross-material\\
        \bottomrule
        \end{tabular}
        }
        \vspace{-1.6em}
    \end{table*}
    
    There are numerous published approaches to liveness detection, which can be classified as hardware-based, software-based or a combination of both. Hardware based methods use a number of additional sensors to gain further insight into the liveness of the presented fingerprint~\cite{baldisserra2006fake, lapsley1998anti, engelsma2018raspireader}. Similarly, a few sensing technologies are inherently well suited for liveness detection and have been used for fingerprint PAD, such as the multispectral Lumidigm sensor or OCT based senors~\cite{chugh2019oct}. On the other hand, software-based solutions use only the information contained in the captured fingerprint image (or a sequence of images) to classify a fingerprint as bonafide or PA~\cite{marcialis2010analysis, marasco2012combining, ghiani2012fingerprint, ghiani2013fingerprint, nogueira2016fingerprint, pala2017deep, chugh2018fingerprint}. Of the existing software solutions, convolutional neural network (CNN) approaches have shown the best performance on the respective genuine vs. PA benchmark datasets. However, it has been shown that the spoof detection error rates of these approaches suffer up to a three fold increase when applied to datasets containing spoof materials not seen during training, denoted as \textit{cross-material generalization}~\cite{marasco2011robustness, tan2010effect}.
    
    Some published studies aimed at reducing the performance gap due to cross material evaluations are summarized in Table \ref{tab:prior_work}. A similar performance gap exists for \textit{cross-sensor generalization}, in which presentation attack algorithms are applied to fingerprint images captured on new fingerprint sensor devices that were not seen during training. One explanation for the challenge of cross-sensor generalization is the different textural characteristics in the fingerprint images from different sensors (Figure~\ref{fig:6_sensors}). This discrepancy in the representation performance between the \emph{seen source domain} and the \emph{unseen target domain} has been referred to as the ``domain gap" in the growing literature of deep neural network models applied for representational learning~\cite{bengio2013representation}. The cross-sensor evaluation can be considered as two separate cases: (i) all sensors in the evaluation employ the same sensing technology, e.g., all optical FTIR, and (ii) the sensors may vary in the underlying sensing mechanisms used, e.g., optical direct-view vs. capacitive.
    
    
    \begin{figure*}[t]
        \centering
        \includegraphics[width=\textwidth]{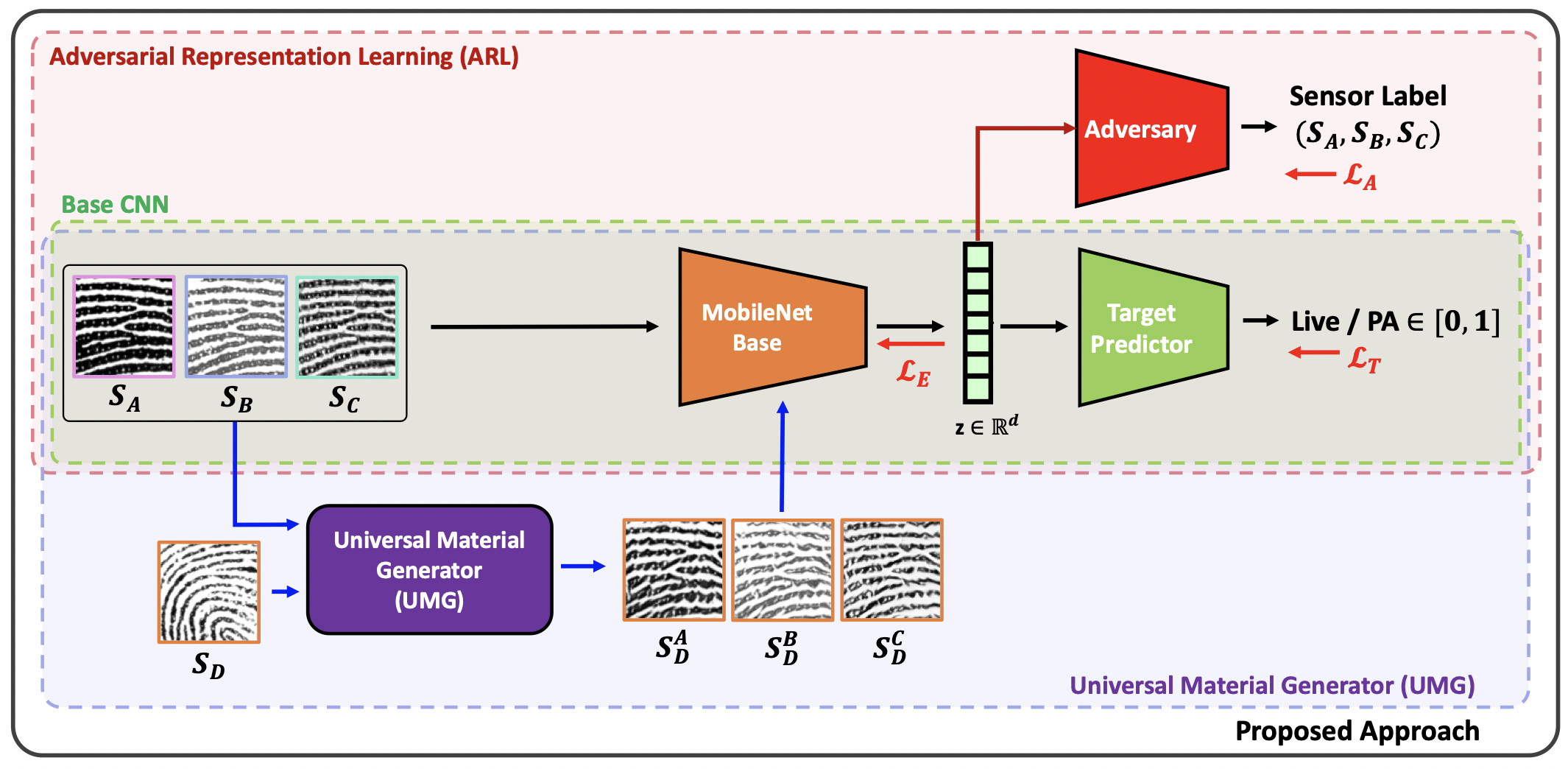}%
        \caption{Overview of the network architecture for the proposed UMG + ARL approach for live vs. presentation attack (PA) detection. $S_A, S_B, S_C,$ and $S_D$ represent fingerprint images from four different fingerprint sensors. $\mathcal{L}_T$ denotes a cross-entropy loss on the target prediction, $\mathcal{L}_A$ denotes a cross-entropy loss on the sensor label prediction, and $\mathcal{L}_E$ denotes the loss propagated to the encoder.}
        \label{fig:overview}
        \vspace{-1.5em}
    \end{figure*}
    
    In this work, we aim to improve the fingerprint presentation attack detection generalization across novel spoof materials and fingerprint sensing devices\footnote{Generally, fingerprint sensor refers to the fingerprint sensing mechanism (e.g., camera and prism for FTIR optical, direct-view camera, thermal measurement device, etc.) and fingerprint reader refers to the entire process of converting a physical fingerprint into a digital image. In this work, similar to the literature, we use these two terms interchangeably.}. Our approach builds off any existing CNN-based architecture trained for fingerprint liveliness detection combined with cross material spoof generalization using a style transfer network wrapper. We also incorporate adversarial representation learning (ARL) in deep neural networks (DNN) to learn sensor and material invariant representations for presentation attack detection.
    
    The main contributions of this study are enumerated below:
    \begin{enumerate}
        \item A robust PAD solution with improved cross-material and cross-sensor generalization performance.
        \item Our solution can be built on top of any CNN-based fingerprint PAD solution for cross-sensor and cross-material spoof generalization using adversarial representational learning.
        \item Experimental evaluation of the proposed approach on publicly available datasets LivDet 2015, LivDet 2017, and MSU-FPAD. Our approach is shown to improve the cross-sensor (cross-material) generalization performance from a TDR of $88.36 \%$ ($78.76 \%$) to a TDR of $92.94 \%$ ($87.86 \%$) at a FDR of $0.2 \%$.
        \item Feature space analysis of cross-sensor domain separation of the embedded representations prior to and following adversarial representation learning.
        \item Detailed discussion of the challenges and techniques involved in applying deep-adversarial representation learning for spoof detection.
    \end{enumerate}
\section{Related Work}
    In this section we briefly discuss the preliminaries of domain adaptation and domain generalization in the context of machine learning. Csurka provides a more in-depth review of domain adaptation~\cite{csurka2017domain}. Similarly, Wang and Deng provide a specific survey of the recent deep domain adaptation methods~\cite{wang2018deep}. We also describe adversarial representation learning (ARL) as it is applied to the tasks of domain adaptation and domain generalization.
    
    \subsection{Domain Adaptation and Domain Generalization}
        A domain refers to a probability distribution over which data examples are drawn from. In this context, domain adaptation and domain generalization are approaches to machine learning aimed at minimizing the performance gap between training data examples from a seen ``source" domain and testing data from a related, but different ``target" domain. Therefore, domain adaptation and domain generalization are applied to situations in which the training and testing data points are not both independently and identically sampled from the same distribution. While domain adaptation involves training on labeled examples from the source domain and unlabeled data from the target domain, domain generalization assumes no access to labeled or unlabeled data examples from the target domain.
        
    \subsection{Adversarial Representation Learning (ARL)}
        Adversarial representation learning is a machine learning technique that can be applied to both domain adaptation and domain generalization. Adversarial representation learning has been applied in DNN architectures to extract discriminative representations for a given target prediction task (say face recognition), while obfuscating some undesired attributes present in the data (say gender information)~\cite{edwards2015censoring, ganin2016domain, tzeng2017adversarial, zhang2018mitigating}.
        
        The general setup of ARL involves (i) an encoder network, (ii) a target prediction network, and (iii) an adversary network. The encoder network aims to extract a latent representation ($\mathbf{z}$) that is not only informative for the target prediction task ($t$), but also does not leak any information for the sensitive task ($s$). Meanwhile, the adversary network is tasked with extracting the sensitive information from the encoded latent representation. The entire network is trained in a minimax game similar to the generative adversarial networks introduced by Goodfellow et al.~\cite{goodfellow2014generative}.
        
        In Xie et al., the parameters of the adversary network are optimized to maximize the likelihood of the sensitive label prediction, whereas the encoder is trained to maximize the likelihood of the target task, while minimizing the likelihood of the sensitive task~\cite{xie2017controllable}. In contrast, our proposed work is more aligned with the approach proposed by Roy and Bodetti~\cite{roy2019mitigating}, where the adversary network is optimized to maximize the likelihood of the sensitive label prediction from the latent representation and the encoder is trained to maximize the entropy of the sensitive label prediction. In this manner, the base network is encouraged to encode a representation that aims to confuse the sensitive label prediction such that the adversary predicts equal probabilities (maximum entropy) for all classes of the sensitive label. 
        

\section{Proposed Approach}
    Our proposed approach is multifaceted and combines ideas from style transfer, which was previously applied for spoof detection, and adversarial representation learning to improve the generalization performance of PAD across different fingerprint sensing devices. An overview of the approach which highlights each of the individual components is shown in Figure~\ref{fig:overview}. Here we introduce each individual component and later discuss the generalization performance improvement achieved with the incorporation of each technique leading up to the final approach.
    
    \subsection{Base CNN}
        What we refer to as the \textit{base CNN} approach is a convolutional neural network (CNN) trained on $96$ x $96$ aligned minutiae-centered patches for classifying a given fingerprint impression as live or spoof. As was shown by Chugh and Jain~\cite{chugh2018fingerprint}, utilizing minutiae patches, as opposed to whole images, overcomes the difficulty in processing fingerprint images of different sizes, provides large amounts of training examples suitable to training deep CNN architectures, and encourages the network to learn local textural cues to robustly separate bonafide from fake fingerprints. This base CNN approach is illustrated in Figure \ref{fig:overview} as the box enclosed by the green line.
        
        The specific architecture of the CNN model employed is the MobileNet-v1 model~\cite{howard2017mobilenets} (the same as in \cite{chugh2018fingerprint})\footnote{Any other CNN-based approach other than~\cite{chugh2018fingerprint} can be used instead.}, where the final 1000-unit softmax layer is replaced with a 2-unit softmax layer suitable for the two-class problem of live vs. spoof. The network is trained from scratch with an RMSProp optimizer at a batch size of 64. During training, data augmentation tools of random distorted cropping, horizontal flipping and random brightness were employed to encourage robustness to overfitting to minute variations of the input images.
        

    \subsection{Adversarial Representational Learning (ARL)}
        ARL is an approach to domain generalization that does not require any knowledge of the unseen target domain, yet aims to learn a generalized and robust feature representation for both source and target domains. The goal of the \textit{ARL} approach is to encourage an encoding network to learn a representation that is invariant to which sensor generated the input fingerprint images (sensitive label), while accurately predicting live vs. PA (target label).
        
        In this setup, the encoder network is represented as a deterministic function, $\mathbf{z} = E(\mathbf{x};\mathbf{\theta_{E}})$, the target prediction network estimates the conditional distribution $p(t|\mathbf{x})$ through $q_{T}(t|\mathbf{z};\mathbf{\theta_{T}})$, and the adversary network estimates the conditional distribution $p(s|\mathbf{x})$ through $q_{A}(s|\mathbf{z};\mathbf{\theta_{A}})$; where $\mathbf{x}$ denotes the input fingerprint image, $p(t|\mathbf{x})$ and $p(s|\mathbf{x})$ represent the probabilities of ground truth target and sensitive labels $t$ and $s$, respectively.
        
        To learn this sensor-invariant representation, the adversary network is trained to maximize the likelihood of predicting which sensor generated the input fingerprint image from the encoded representation. The parameters, $\mathbf{\theta_{A}}$, of the adversary network are updated to minimize the loss defined in equation~\ref{eq:L_A}. The output of the adversary network is used to encourage the encoder to produce a representation that obfuscates the sensitive class labels by penalizing the parameters of the encoder, $\mathbf{\theta_{E}}$, to minimize the loss in equation~\ref{eq:L_E}, where $\alpha$ is a hyper-parameter that allows for a trade-off between obfuscation of the sensitive label and prediction of the target label. Meanwhile, to accurately predict live vs. PA, the parameters of target prediction network, $\mathbf{\theta_{T}}$, are optimized to minimize the loss in equation~\ref{eq:L_T}. The ARL approach is shown in Figure \ref{fig:overview} by the box enclosed by the red line.
        
        \small{
        {\setlength{\mathindent}{0cm}
        \begin{equation}
            \label{eq:L_A}
            \mathcal{L}_{A} = \mathbb{E}_{\mathbf{x},s}[-\log q_{A}(s|E(\mathbf{x};\mathbf{\theta_{E}});\mathbf{\theta_{A}})]
        \end{equation}
        \vspace{-1.5em}
        \begin{multline}
            \label{eq:L_E}
            \mathcal{L}_{E} = \mathbb{E}_{\mathbf{x},t}[-\log q_{T}(t|E(\mathbf{x};\mathbf{\theta_{E}});\mathbf{\theta_{T}})] \\ + \alpha \mathbb{E}_{\mathbf{x}}[\sum^{m}_{i=1}q_{A}(s_{i}|E(\mathbf{x};\mathbf{\theta_{E}});\mathbf{\theta_{A}})\log q_{A}(s_{i}|E(\mathbf{x};\mathbf{\theta_{E}});\mathbf{\theta_{A}})]
        \end{multline}
        \vspace{-1.5em}
        \begin{equation}
            \label{eq:L_T}
            \mathcal{L}_{T} = \mathbb{E}_{\mathbf{x},t}[-\log q_{T}(t|E(\mathbf{x};\mathbf{\theta_{E}});\mathbf{\theta_{T}})]
        \end{equation}
        }
        \vspace{-1.5em}
        }

    \subsection{Na\"{i}ve}
        A simple approach to cross-sensor generalization is one in which we assume access to a limited number of training examples ($100$ live and PA fingerprint images) from the target sensor that we include during training, which doesn't require collecting extensive amounts of data from the target domain. This is a reasonable assumption in the case of cross-sensor generalization, where we have access to the sensing device on which the system will be deployed. This is in contrast to generalization to unknown spoof materials, where we cannot assume any prior knowledge of the unknown target materials. We denote this method as the \textit{na\"{i}ve} approach to cross-sensor spoof detection as it does not require any changes to the system architecture.
        
    \subsection{Na\"{i}ve + ARL}
        We combine the na\"{i}ve approach with ARL to take advantage of the benefits of each separate approach. By exposing the network to the textural characteristics inherent to the small number of target sensor images during training, the goal is that the network will better learn a mapping from images to representations for each sensor domain. Furthermore, by incorporating the adversary during training to learn a sensor-invariant representation, we aim to overcome the apparent imbalance in the number of training examples from source and target sensors.

    \subsection{Universal Material Generator (UMG)}
        The final technique that we incorporate is a style transfer approach, coupled on top of the na\"{i}ve approach, to augment the training data from the target sensor. The specific style transfer network we use is the Universal Material Generator (UMG) proposed in~\cite{chugh2019fingerprint} that inputs source and target domain minutiae patches and produces a large amount of synthetic training images in the target sensor domain. UMG achieves this by learning a mapping from the style of the source domain image patches to the style of the target domain image patches. Concretely, the UMG separates the content information, i.e, the fingerprint ridge structure, and the style, i.e, textural information, of a given fingerprint minutiae patch and produces a synthetic image that has the content of the source domain and the style of the target domain. An overview of the \textit{UMG} approach is shown as the box enclosed by the blue line in Figure \ref{fig:overview}.
    
    \subsection{UMG + ARL (Proposed Approach)}
        The proposed approach applies ARL with the UMG style transfer wrapper to further improve generalization performance. An illustration of the \textit{ARL + UMG} approach is illustrated in Figure~\ref{fig:overview} as everything enclosed by the box formed by the solid, black line. Like the na\"{i}ve approach, this method inherently assumes knowledge of a limited set of examples from the target domain sensor. Specifically, we assume $100$ live and $100$ PA images from the target sensor. From this small set of images from the target sensor, we produce a much larger set of synthetic images in the target domain using the UMG wrapper to transfer the style of the target domain to the content of the source domain training images.
        
        The advantage of this approach is that we leverage the ability of the UMG wrapper to ensure a balanced dataset from all sensors (source and target), which we combine with ARL that forces the network to learn a sensor-invariant representation. In the following section, we demonstrate the performance gains over the previous approaches and show that the UMG coupled with ARL achieves the new state-of-the-art in cross-sensor and cross-material generalization of fingerprint PAD.

\section{Evaluation Procedure}
    In this section we describe the experimental protocol of the various experiments carried out in this study, the datasets involved in each experiment, and the implementation details of the UMG + ARL approach. 
    \subsection{Experimental Protocol}
        To evaluate cross-sensor PAD performance, we adopt the leave-one-out protocol where one sensor is set aside for testing and the network is trained on data from all remaining sensors. To analyze separately the cross-sensor performance and the cross-material performance, we segment our evaluation to include the case where all materials during testing were included during training (cross-sensor only) and the case where no materials during training were seen in testing (cross-sensor and cross-material).
        
        \begin{table*}
            \renewcommand{\arraystretch}{1.3}
            \caption{Summary of the 2015 and 2017 Liveness Detection (LivDet) Datasets.}
            \label{tab:livdet}
            \centering
            \scriptsize 
            {
            \begin{tabular}{|p{0.11\textwidth}||p{0.1\textwidth}|p{0.1\textwidth}|p{0.1\textwidth}|p{0.1\textwidth}||p{0.09\textwidth}|p{0.09\textwidth}|p{0.09\textwidth}|}
            \hline
            Dataset & \multicolumn{4}{c||}{LivDet 2015} & \multicolumn{3}{c|}{LivDet 2017}\\
            \hline
            Fingerprint Reader & Green Bit & Biometrika & Digital Persona & CrossMatch & Green Bit & Orcanthus & Digital Persona\\
            \hline
            Model & DactyScan26 & HiScan-PRO & U.are.U 5160 & L Scan Guardian & Dacty Scan 84C & Cerits2 Image & U.are.U 5160\\
            \hline
            Image Size & 500 x 500 & 1000 x 1000 & 252 x 324 & 640 x 480 & 500 x 500 & 300 x $n^\dagger$ & 252 x 324\\
            \hline
            Resolution (dpi) & 500 & 1000 & 500 & 500 & 569 & 500 & 500\\
            \hline
            \#Live Images \newline Train / Test & 1000 / 1000 & 1000 / 1000 & 1000 / 1000 & 1510 / 1500 & 1000 / 1700 & 1000 / 1700 & 1000 / 1692\\
            \hline
            \#Spoof Images \newline Train / Test & 1000 / 1500 & 1000 /1500 & 1000 / 1500 & 1473 / 1448 & 1200 / 2040 & 1180$^{*}$/ 2018 & 1199 / 2028\\
            \hline
            Spoof Materials & \multicolumn{3}{p{0.3\textwidth}|}{Ecoflex, Gelatine, Latex, Wood Glue, Liquid Ecoflex, RTV} & Body Double, Ecoflex, PlayDoh, OOMOO, Gelatin & \multicolumn{3}{p{0.3\textwidth}|}{Wood Glue, Ecoflex, Body Double, Gelatine, Latex, Liquid Ecoflex}\\
            \hline
            \multicolumn{8}{p{0.9\textwidth}}{$^\dagger$ Fingerprint images captured by Orcanthus have a variable height (350 - 450 pixels) depending on the friction ridge content.}\\
            \multicolumn{8}{p{0.95\textwidth}}{$^{*}$ A Set of 20 Latex spoof fingerprints were present in the training data of Orcanthus; which were excluded in our experiments because only Wood Glue, Ecoflex, and Body Double are expected to be in the training dataset.}\\
            \end{tabular}
            }
            \vspace{-1.0em}
        \end{table*}
        
        \begin{table}
            \renewcommand{\arraystretch}{1.3}
            \caption{Summary of the MSU-FPAD Dataset.}
            \label{tab:MSU-FPAD}
            \centering
            \scriptsize 
            {
            \begin{tabular}{|p{0.13\textwidth}||p{0.12\textwidth}|p{0.12\textwidth}|}
            \hline
            Dataset & \multicolumn{2}{c|}{MSU-FPAD}\\
            \hline
            Fingerprint Reader & CrossMatch & Lumidigm\\
            \hline
            Model & Guardian 200 & Venus 302\\
            \hline
            Image Size & 750 x 800 & 400 x 272\\
            \hline
            Resolution (dpi) & 500 & 500\\
            \hline
            \#Live Images \newline Train / Test & 2250 / 2250 & 2250 / 2250\\
            \hline
            \#Spoof Images \newline Train / Test & 3000 / 3000 & 2250 / 2250\\
            \hline
            Spoof Materials & \multicolumn{2}{p{0.2\textwidth}|}{Ecoflex, PlayDoh, 2D Print (Matte Paper), 2D Print (Transparency)}\\
            \hline
            \end{tabular}
            }
            \vspace{-1.0em}
        \end{table}
        
    \subsection{Datasets}
        The data used in the experiments for this paper are from the LivDet 2015, LivDet 2017, and MSU-FPAD datasets, which are summarized in Tables \ref{tab:livdet} and \ref{tab:MSU-FPAD}. The LivDet 2015 dataset consists of four sensors: Biometrika, CrossMatch, Digital Persona, and Green Bit. These sensors are all FTIR optical image capturing devices. We utilize this dataset to evaluate the generalization performance across different fingerprint readers with the same sensing technology. To further evaluate our approach on fingerprint readers with different sensing mechanisms, we experiment on fingerprint data from the Lumidigm sensor of the MSU-FPAD dataset. This sensor uses different sensing technology from the four seen in the LivDet 2015 as it is a  multi-spectral, direct-view capture device. Finally, we incorporate a third dataset, LivDet 2017, which consists of three sensors: Digital Persona, Green Bit, and Orcanthus, where Orcanthus uses thermal-based imaging.
        
    \subsection{Implementation Details}
        The architecture of the encoder in the proposed approach is MobileNet-v1 with the final $1000$-unit softmax layer removed, which is used to encode a latent representation $\mathbf{z}\in\mathbb{R}^{d}$. In our implementation, $d = 1024$. The target predictor is a single fully connected layer of $2$-dimensions (for predicting live vs. PA) with a softmax activation. The adversary network consists of a fully connected layer with a softmax activation of output dimension equal to the number of source sensors in the training dataset, e.g., $3$ in the leave-one-out protocol on the LivDet 2015 dataset.
        
        Training adversarial losses is notoriously difficult and often requires extensive hyper-parameter tuning. For example, it was found advantageous during training to update the parameters, $\mathbf{\theta_{A}}$, of the adversary network five times per every update of the encoder and target predictor. We also explored adjusting the number of hidden layers in the adversary network, but no significant improvements over a single layer network were observed. A grid search was performed over the value of $\alpha$ for selecting the influence of the adversary on updating the parameters, $\mathbf{\theta_{E}}$, of the encoder, and the optimal parameter value of $\alpha = 0.1$ was selected (See Eq.~\ref{eq:L_E}).
    
\section{Experimental Results}

    \begin{table*}[t]
        \renewcommand{\arraystretch}{1.3}
        \caption{Cross-Sensor Generalization Performance (TDR (\%) @ FDR = 0.2 \%)$^{\dagger}$ with Leave-One-Out Method on LivDet 2015 Dataset with Materials Common to Training and Testing, i.e., Excluding Cross-Materials$^{\ddagger}$. Bio = Biometrika, CM = CrossMatch, DP = Digital Persona, and GB = GreenBit.}
        \label{tab:cross-sensor-results}
        \scriptsize
        {
        \begin{tabular}{C{0.1\textwidth}|*{4}{C{0.08\textwidth}C{0.035\textwidth}|}*{2}{C{0.085\textwidth}}}
        &Source$^{*}$\newline CM, DP, GB &Target$^{\star}$\newline Bio &Source\newline Bio, DP, GB &Target\newline CM &Source\newline Bio, CM, GB &Target\newline DP &Source\newline Bio, CM, DP &Target\newline GB &\textbf{Source\newline Mean $\pm$ s.d.} &\textbf{Target\newline Mean $\pm$ s.d.}\\
        \toprule
        Base CNN~\cite{chugh2018fingerprint} & 90.34 & 75.16 & 88.20 & 3.33 & 98.40 & 10.76 & 92.82 & 70.74 & 92.44 $\pm$ 4.40 & 40.00 $\pm$ 38.21\\
        \hline
        ARL & 93.44 & 80.51 & 91.03 & 2.11 & 98.73 & 11.74 & 92.04 & 64.74 & \textbf{93.81 $\pm$ 3.43} & 39.78 $\pm$ 38.67\\
        \hline
        Na\"{i}ve & 87.74 & 84.80 & 88.23 & 97.37 & 96.96 & 59.13 & 88.08 & 90.68 & 90.25 $\pm$ 4.48 & 83.00 $\pm$ 16.72\\
        \hline
        UMG~\cite{chugh2019fingerprint} & 89.10 & 94.33 & 84.28 & 90.70 & 96.39 & 71.85 & 78.14 & 96.57 & 86.98 $\pm$ 7.71 & 88.36 $\pm$ 11.27\\
        \hline
        Na\"{i}ve + ARL & 90.18 & 91.86 & 87.87 & 98.95 & 94.21 & 52.07 & 89.15 & 83.92 & 90.35 $\pm$ 2.74 & 81.70 $\pm$ 20.69\\
        \hline
        UMG + ARL & 88.98 & 92.83 & 88.48 & 97.54 & 96.18 & 87.61 & 86.88 & 93.78 & 90.13 $\pm$ 4.13 & \textbf{92.94 $\pm$ 4.09}\\
        \bottomrule
        \multicolumn{11}{l}{$\dagger$ We use FDR = 0.2 \% because this is the stringent metric being used by the IARPA Odin program. Due to space limits, it is challenging to show the complete Receiver}\\
        \multicolumn{11}{l}{Operating Curve (ROC) or Detection Error Tradeoff (DET) curve.}\\
        \multicolumn{11}{l}{$\ddagger$ Liquid Ecoflex and RTV materials were excluded from the testing sets of Green Bit, Biometrika, and Digital Persona. Body Double, Playdoh, and OOMOO were excluded}\\
        \multicolumn{11}{l}{from the testing set of CrossMatch.}\\
        \multicolumn{11}{l}{$*$ Sensors included in the training set (source)}\\
        \multicolumn{11}{l}{$\star$ Sensors included in the test set (target)}\\
        \end{tabular}
        }
        \vspace{-1.0em}
    \end{table*}

    \begin{table*}[t]
        \renewcommand{\arraystretch}{1.3}
        \caption{Cross-Sensor and Cross-Material Generalization Performance (TDR (\%) @ FDR = 0.2 \%) with Leave-One-Out Method on LivDet 2015 Dataset with Materials Exclusive to the Testing Datasets, i.e., Cross-Material Only. Bio = Biometrika, CM = CrossMatch, DP = Digital Persona, and GB = GreenBit.}
        \label{tab:cross-sensor-and-cross-material-results}
        \scriptsize
        {
        \begin{tabular}{C{0.1\textwidth}|*{4}{C{0.08\textwidth}C{0.035\textwidth}|}*{2}{C{0.085\textwidth}}}
         &Source\newline CM, DP, GB &Target\newline Bio &Source\newline Bio, DP, GB &Target\newline CM &Source\newline Bio, CM, GB &Target\newline DP &Source\newline Bio, CM, DP &Target\newline GB &\textbf{Source\newline Mean $\pm$ s.d.} &\textbf{Target\newline Mean $\pm$ s.d.}\\
        \toprule
        Base CNN~\cite{chugh2018fingerprint} & 90.34 & 63.92 & 88.20 & 4.46 & 98.40 & 11.39 & 92.82 & 72.39 & 92.44 $\pm$ 4.40 & 38.04 $\pm$ 35.06\\
        \hline
        ARL & 92.78 & 72.58 & 91.03 & 6.06 & 98.73 & 13.08 & 92.04 & 49.69 & \textbf{93.65 $\pm$ 3.47} & 35.35 $\pm$ 31.33\\
        \hline
        Na\"{i}ve & 87.74 & 77.11 & 88.23 & 96.80 & 96.96 & 42.62 & 88.08 & 85.69 & 90.25 $\pm$ 4.48 & 75.56 $\pm$ 23.39\\
        \hline
        UMG~\cite{chugh2019fingerprint} & 89.10 & 87.01 & 84.28 & 81.37 & 96.39 & 54.43 & 78.14 & 92.23 & 86.98 $\pm$ 7.71 & 78.76 $\pm$ 16.82\\
        \hline
        Na\"{i}ve + ARL & 90.18 & 86.19 & 87.87 & 97.45 & 94.21 & 35.65 & 82.51 & 65.44 & 88.69 $\pm$ 4.88 & 71.18 $\pm$ 27.15\\
        \hline
        UMG + ARL & 89.31 & 89.07 & 88.48 & 92.69 & 96.18 & 78.69 & 86.88 & 91.00 & 90.21 $\pm$ 4.10 & \textbf{87.86 $\pm$ 6.29}\\
        \bottomrule
        \end{tabular}
        }
        \vspace{-1.0em}
    \end{table*}
    
    Here we present the results of each experiment to evaluate the cross-sensor and cross-material generalization performance of the proposed approach. This section is divided into several parts to facilitate an in-depth analysis of the generalization performance of the algorithm to each of the following cases: cross-sensor, cross-material, and cross-sensing technology. A discussion on the effect of varying the number of assumed target domain images is included in section 5.4. We conclude this section with an analysis of the deep feature space prior to and following the application of the proposed methodology for fingerprint spoof generalization. The feature space analysis is conducted utilizing a 2-dimensional t-Distributed Stochastic Neighbor Embedding (t-SNE) visualization~\cite{maaten2008visualizing}.
    
    There has not been much prior work aimed specifically at improving cross-sensor generalization of fingerprint PAD; nonetheless, there are a few cross-sensor performance results reported in the literature. Chugh and Jain report the cross-sensor performance of Fingerprint Spoof Buster, which shares the same architecture of our base encoder model~\cite{chugh2018fingerprint}. Therefore, in the following sections we compare our performance against that of Fingerprint Spoof Buster as the Base CNN model. Furthermore, Chugh and Jain report cross-sensor results in their work toward improving cross-material generalization with the introduction of their UMG network wrapper~\cite{chugh2019fingerprint}. For comparison with this approach, we refer to their work as the UMG approach in Tables \ref{tab:cross-sensor-results} and \ref{tab:cross-sensor-and-cross-material-results} of this section.
    
    \subsection{Cross-Sensor Performance}
        To evaluate cross-sensor generalization we utilize the LivDet 2015 dataset which consists of four different FTIR optical fingerprint imaging devices and we apply a leave-one-out strategy where the algorithm is trained on only three of the four sensors at a time. We then compare the performance on a test set of data from these three sensors included in the training to the performance on a test set consisting of data from the remaining sensor. We repeat this procedure for all four combinations of sensors and report the results in Table \ref{tab:cross-sensor-results}.
        
        To separate out the cross-sensor generalization performance from the related task of cross-material generalization, we first remove all the non-overlapping materials between the testing dataset of the target sensor and the training datasets of the three source sensors. For this experiment, Liquid Ecoflex and RTV materials were excluded from the testing sets when Green Bit, Biometrika, and Digital Persona were the target sensors; whereas, Body Double, Playdoh, and OOMOO were excluded from the testing set with CrossMatch as the target sensor.
        
        As shown in Table~\ref{tab:cross-sensor-results}, the proposed approach of UMG + ARL increases the average cross-sensor generalization in terms of True Detection Rate (TDR) at a False Detection Rate (FDR) of $0.2 \%$\footnote{We consider this metric to be more representative of actual use cases as opposed to EER and ACE. Space limitation does not allow us to show the full ROC curve.} from $88.36 \%$ to $92.94 \%$ over the UMG only method. The proposed approach also maintains higher performance (TDR = $90.13 \%$) on the source domain sensors compared to the UMG only approach (TDR = $86.98 \%$). Lastly, we note that the standard deviation (s.d.) across the four experiments of cross-sensor generalization on the LivDet 2015 dataset is significantly reduced for the UMG + ARL method ($11.27 \%$ to $4.09 \%$), in comparison UMG only, indicating the robustness of the proposed approach.
        
        For completeness, we include an evaluation of using an additional CNN architecture, Resnet-v1-50\footnote{Resnet-v1-50 was chosen since the authors of other SOTA fingerprint PAD algorithms were not willing to share their code and we found the details of their reported implementations insufficient for reproducing for a fair evaluation.}~\cite{he2016deep}, as the base encoder to demonstrate the generality of the proposed approach. In Table~\ref{tab:resnet_results}, we report the performance with ResNet-v1-50 as the Base CNN model on LivDet 2015 with leaving Biometrika out as the target sensor. We see that the performance improvement is consistent for both Base CNN models, supporting the generality of the approach to any existing CNN architecture trained for fingerprint spoof detection. In the remaining experiments, we continue to report results for only Spoof Buster as the Base CNN model.
        
        \begin{table}
            \renewcommand{\arraystretch}{1.3}
            \caption{Cross-Sensor Generalization Performance (TDR (\%) @ FDR = 0.2 \%) on Leave-Out Biometrika (LivDet 2015) using Resnet-v1-50 as the Base CNN Model. Bio = Biometrika, CM = CrossMatch, DP = Digital Persona, and GB = GreenBit.}
            \label{tab:resnet_results}
            \scriptsize
            {
            \begin{tabular}{C{0.1\textwidth}C{0.15\textwidth}C{0.15\textwidth}}
             & \hfil Source \newline \hfil CM, DP, GB & \hfil Target \newline \hfil Bio\\
            \toprule
            Base CNN~\cite{he2016deep} & 65.29 & 76.02\\
            \hline
            ARL & 72.72 & 72.27\\
            \hline
            Na\"{i}ve & 73.55 & 90.79\\
            \hline
            UMG~\cite{chugh2019fingerprint} & 72.76 & 91.76\\
            \hline
            Na\"{i}ve + ARL & 73.05 & 92.18\\
            \hline
            UMG + ARL & \textbf{75.94} & \textbf{92.83}\\
            \bottomrule
            \end{tabular}
            }
            \vspace{-1.6em}
        \end{table}
        
    \subsection{Cross-Sensor and Cross-Material Performance}
        We now compare the performance of each solution on the cross-sensor and cross-material experiment by following the same procedure as the cross-sensor experiment, while including only materials exclusive to the test datasets of LivDet 2015. Even though our system was trained to adversarially learn a sensor-invariant representation, we report the results of including unseen materials to evaluate whether we automatically obtain the added benefit of cross-material generalization (Table~\ref{tab:cross-sensor-and-cross-material-results}).
        
        The results of Table~\ref{tab:cross-sensor-and-cross-material-results} agree with the results of the cross-sensor only experiment shown previously; however, we note small performance declines due to the evaluation on only unknown spoof materials. Specifically, the average TDR at a FDR of $0.2 \%$ of the proposed approach decreased from $92.38 \%$ for cross-sensor only to $87.86 \%$ for cross-sensor and cross-material generalization on the target sensor. However, we notice that the performance degradation of the UMG + ARL method is less than the drop in performance of the UMG only approach, which further demonstrates the generalization benefits of incorporating ARL for fingerprint PAD. It seems that learning an invariance to the textural differences between different sensors also encourages an invariance to the textural differences between different spoof materials.
        
    \subsection{Cross-Sensing Technology Performance}
        In this section, we expand our analysis to include generalization across different fingerprint sensing mechanisms, where the sensing technology of the source fingerprint readers during training is different from the target test reader. For the first experiment we incorporate the data from the Lumidigm multispectral sensor of the MSU-FPAD database as the test sensor and the four FTIR optical sensors of LivDet 2015 as our training sensors. In this experiment we do not control for unknown materials between training and test sets, thus we could consider the evaluation as a combination of cross-sensor, cross-material, and cross-sensing technology. The results show that UMG + ARL achieves the highest generalization TDR of $88.60 \%$ on the target domain sensor (Figure~\ref{tab:lumidigm_results}).
        
        To further evaluate the generalization performance of the proposed UMG + ARL approach, we repeat the experiments on a third dataset, LivDet 2017, which consists of data from three different sensors: Green Bit (optical FTIR), Digital Persona (optical FTIR), and Orcanthus (thermal). With the inclusion of the Orcanthus sensor as a thermal based technology, we can evaluate cross-sensing technology performance where the underlying imaging technology between the sensors is substantially different. Further, we do not remove unseen material types between the training and testing datasets of LivDet 2017 for this experiment. As shown in Table~\ref{tab:livdet2017_results}, the generalization performance (TDR @ FDR $= 0.2 \%$) on LivDet 2017 improves over the state-of-the-art from $34.80 \%$ to $36.47 \%$. 
        
        \begin{table}
            \renewcommand{\arraystretch}{1.3}
            \caption{Cross-Sensing Technology Generalization Performance (TDR (\%) @ FDR = 0.2 \%) with Four Sensors of LivDet 2015 Dataset Included During Training and Lumidigm from the MSU-FPAD Dataset Left Out For Testing. Bio = Biometrika, CM = CrossMatch, DP = Digital Persona, GB = GreenBit, and Lum = Lumidigm.}
            \label{tab:lumidigm_results}
            \scriptsize
            {
            \begin{tabular}{C{0.1\textwidth}*{2}{C{0.15\textwidth}}}
             & \hfil Source\newline \hfil Bio, CM, DP, GB & \hfil Target\newline \hfil Lum\\
            \toprule
            Base CNN~\cite{chugh2018fingerprint} & \textbf{90.40} & 0.60\\
            \hline
            ARL & 87.41 & 3.00\\
            \hline
            Na\"{i}ve & 63.54 & 61.27\\
            \hline
            UMG~\cite{chugh2019fingerprint} & 88.24 & 80.60\\
            \hline
            Na\"{i}ve + ARL & 87.22 & 84.93\\
            \hline
            UMG + ARL & 88.45 & \textbf{88.60}\\
            \bottomrule
            \end{tabular}
            }
            \vspace{-1.0em}
        \end{table}
        
        \begin{table}
            \renewcommand{\arraystretch}{1.3}
            \caption{Cross-Sensing Technology Generalization Performance (TDR (\%) @ FDR = 0.2 \%) on LivDet 2017 Dataset.}
            \label{tab:livdet2017_results}
            \scriptsize
            {
            \begin{tabular}{C{0.1\textwidth}*{2}{C{0.15\textwidth}}}
             & \hfil \textbf{Source\newline \hfil Mean $\pm$ s.d.} & \hfil \textbf{Target\newline \hfil Mean $\pm$ s.d.}\\
            \toprule
            Base CNN~\cite{chugh2018fingerprint} & 41.43 $\pm$ 5.83 & 4.63 $\pm$ 8.71\\
            \hline
            ARL & 38.92 $\pm$ 6.64 & 7.35 $\pm$ 12.27\\
            \hline
            Na\"{i}ve & 43.90 $\pm$ 7.26 & 27.30 $\pm$ 6.82\\
            \hline
            UMG~\cite{chugh2019fingerprint} & 39.02 $\pm$ 14.71 & 34.80 $\pm$ 4.96\\
            \hline
            Na\"{i}ve + ARL & \textbf{44.63 $\pm$ 15.52} & 30.30 $\pm$ 11.97\\
            \hline
            UMG + ARL & 38.50 $\pm$ 14.63 & \textbf{36.47 $\pm$ 9.86}\\
            \bottomrule
            \end{tabular}
            }
            \vspace{-1.0em}
        \end{table}
        
    \subsection{Varying Number of Target Domain Images}
        To study the effect of varying the number of assumed target domain images available during training, we repeat the experiments in the leave-out Biometrika (LivDet 2015) scenario. Specifically, we run experiments on $50$ and $250$ live and PA training images from the target domain. As shown in Table~\ref{tab:num_images}, increasing the number of target domain images greatly benefits the na\"{i}ve approach, but only marginally affects the UMG + ARL method. Therefore, the benefit of UMG + ARL is most pronounced in cases with limited target domain training examples. In the trade-off between time spent for data collection and performance, the proposed method can significantly help reduce the burden of expensive data collection.
    
        \begin{table}
            \renewcommand{\arraystretch}{1.3}
            \caption{Cross-Sensor Generalization Performance (TDR (\%) @ FDR = 0.2 \%) on Leave-Out Biometrika (LivDet 2015) with Varying Number of Target Sensor Training Images.}
            \label{tab:num_images}
            \scriptsize
            {
            \begin{tabular}{C{0.08\textwidth}*{2}{C{0.07\textwidth}}|*{2}{C{0.07\textwidth}}}
             & \multicolumn{2}{c}{50 Images} & \multicolumn{2}{c}{250 Images}\\
             & Source & Target & Source & Target\\
            \toprule
            Na\"{i}ve & 91.21 & 90.15 & 91.04 & 95.29\\
            \hline
            UMG~\cite{chugh2019fingerprint} & \textbf{93.19} & 90.47 & 91.00 & 89.19\\
            \hline
            Na\"{i}ve + ARL & 85.64 & 91.43 & \textbf{95.50} & \textbf{95.40}\\
            \hline
            UMG + ARL & 90.76 & \textbf{93.25} & 90.71 & 93.04\\
            \bottomrule
            \end{tabular}
            }
            \vspace{-1.6em}
        \end{table}
        
    \subsection{Feature Space Analysis}
\begin{figure}
        \centering
        \subfloat[]{\includegraphics[width=0.23\textwidth]{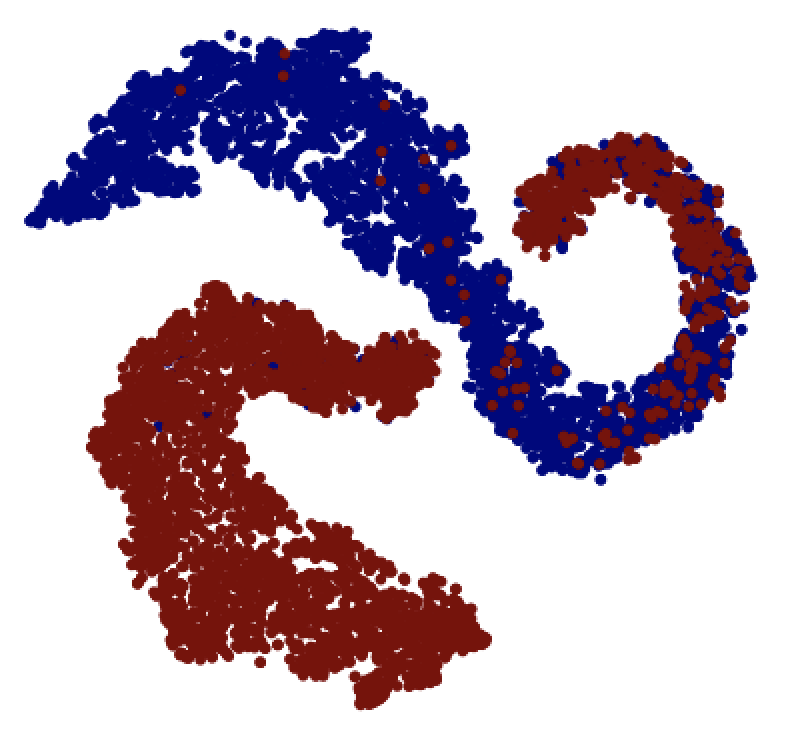}%
        \label{leaveOutCM baseline}}
        \subfloat[]{\includegraphics[width=0.23\textwidth]{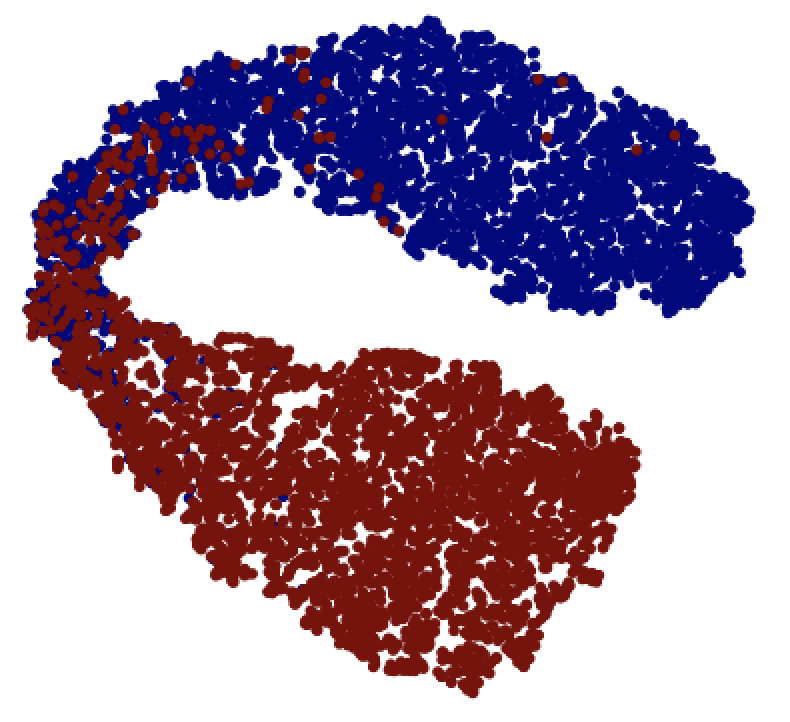}%
        \label{leaveOutCM UMG + ARL}}
        \caption{2-dimensional t-SNE feature embeddings of the target sensor fingerprint minutiae patches for the (a) UMG only and (b) UMG + ARL models trained on the LivDet 2015 dataset with Biometrika, Green Bit, and Digital Persona as the source sensors and CrossMatch as the target sensor. The blue and red dots represent live and spoof minutiae patches of fingerprint impressions captured on the target sensor (CrossMatch), respectively.}
        \label{fig:tsne}
        \vspace{-1.5em}
        \end{figure}

        To explore the benefits of incorporating ARL on top of the UMG only approach, we extract 2-dimensional t-SNE feature embeddings of the live and spoof fingerprint minutiae patches from the final 1024-unit layer of the MobileNet-v1 encoder network, prior to the softmax non-linearity, from the UMG only network and the UMG + ARL network. For brevity, we just show the results of the leave-one-out protocol on the LivDet 2015 dataset with Biometrika, Green Bit, and Digital Persona as the source sensors and CrossMatch as the target sensor. In Figure~\ref{fig:tsne}, we plot these embeddings to analyze the effect of adversarially enforcing the learning of a sensor-invariant representation. Figure~\ref{fig:tsne} (a) shows the separation between live and spoof fingerprint minutiae patch embeddings of the UMG only network for minutiae patches from the target sensor, i.e., CrossMatch, whereas (b) shows the separation of the embeddings produced by the UMG + ARL approach. We can see that the proposed method provides noticeably better separation between the live and fingerprint spoof patches, resulting in the improved PAD performance.
        
\section{Conclusion}
    Diverse and sophisticated presentation attacks pose a threat to the effectiveness of fingerprint recognition systems for reliable authentication and security. Previous PAD algorithms have demonstrated success in scenarios for which significant training data of bonafide and spoof fingerprint images are available, but are not robust to generalize well to novel spoof materials unseen during training. Additionally, previous fingerprint PAD solutions are not generalizable across different fingerprint readers, meaning that a PAD algorithm trained on a specific fingerprint reader will not perform well when applied to different fingerprint sensing devices. 

    The proposed approach towards fingerprint PAD demonstrates an improvement over the state-of-the-art, in terms of true detection rate (TDR) at a false detection rate (FDR) of $0.2 \%$, on cross-sensor and cross-material generalization. In particular, incorporating adversarial representation learning with the Universal Material Generator (UMG) improves the cross-sensor generalization performance from a TDR of $88.36 \pm 11.27 \%$ to $92.94 \pm 4.09 \%$ on the LivDet 2015 dataset, while maintaining higher performance on the sensors seen during training. Further, including cross-materials with the cross-sensor evaluation leads to an improvement of $78.76 \pm 16.82 \%$ to $87.86 \pm 6.29 \%$. Lastly, experiments involving cross-sensor, cross-material, and cross-sensing technology show average improvements of $80.60 \%$ to $88.60 \%$ and $34.80 \%$ to $36.47 \%$ with the proposed approach over state-of-the-art, on the MSU-FPAD and LivDet 2017 datasets, respectively.


\section{Acknowledgment}
    This research is based upon work supported in part by the Office of the Director of National Intelligence (ODNI), Intelligence Advanced Research Projects Activity (IARPA), via IARPA R\&D Contract No. 2017 - 17020200004. The views and conclusions contained herein are those of the authors and should not be interpreted as necessarily representing the official policies, either expressed or implied, of ODNI, IARPA, or the U.S. Government. The U.S. Government is authorized to reproduce and distribute reprints for governmental purposes notwithstanding any copyright annotation therein.

{\small
\bibliographystyle{ieee}
\bibliography{citations}
}

\end{document}